\documentclass[conference]{IEEEtran}
\IEEEoverridecommandlockouts

\usepackage{cite}
\usepackage{amsmath,amssymb,amsfonts}
\usepackage{algorithmic}
\usepackage{graphicx}
\usepackage{textcomp}
\usepackage{xcolor}
\def\BibTeX{{\rm B\kern-.05em{\sc i\kern-.025em b}\kern-.08em
    T\kern-.1667em\lower.7ex\hbox{E}\kern-.125emX}}
\begin{document}

\title{Towards Continuous Skin Sympathetic Nerve
Activity Monitoring: Removing Muscle Noise}

\author{
\IEEEauthorblockN{
Farnoush Baghestani, 
Mahdi Pirayesh Shirazi Nejad, 
Youngsun Kong, 
Ki H. Chon, \textit{Fellow, IEEE}
}
\IEEEauthorblockA{
\textit{Dept. of Biomedical Engineering}, 
\textit{University of Connecticut}, 
Storrs, USA \\
}
}

\maketitle

\begin{abstract}
Continuous monitoring of non-invasive skin sympathetic nerve activity (SKNA) holds promise for understanding the sympathetic nervous system (SNS) dynamics in various physiological and pathological conditions. However, muscle noise artifacts present a challenge in accurate SKNA analysis, particularly in real-life scenarios. This study proposes a deep convolutional neural network (CNN) approach to detect and remove muscle noise from SKNA recordings obtained via ECG electrodes. Twelve healthy participants underwent controlled experimental protocols involving cognitive stress induction and voluntary muscle movements, while collecting SKNA data. Power spectral analysis revealed significant muscle noise interference within the SKNA frequency band ($\bold{500–1000}$ Hz). A 2D CNN model was trained on the spectrograms of the data segments to classify them into baseline, stress-induced SKNA, and muscle noise-contaminated periods, achieving an average accuracy of $\bold{89.85}\%$ across all subjects. Our findings underscore the importance of addressing muscle noise for accurate SKNA monitoring, advancing towards wearable SKNA sensors for real-world applications.
\end{abstract}


\section{Introduction}
The sympathetic nervous system (SNS) exerts diverse cardiovascular effects, such as accelerating heart rate, enhancing cardiac contractility, decreasing venous capacitance, and peripheral vasoconstriction \cite{sinski2006study}. In heart failure, heightened sympathetic activity triggers detrimental effects that exacerbate the disease, including adverse remodeling, changes in the beta-adrenergic receptor system, and abnormalities in skeletal muscle \cite{floras2009sympathetic}. Therefore, studying the SNS is critical for understanding cardiovascular regulation.

Chronic stress can exert adverse effects on both the cardiovascular and nervous systems, manifesting in various symptoms. Elevated levels of distress and anxiety can escalate the likelihood of heart disease, sudden cardiac arrest, and stroke, with the SNS being a pivotal player in initiating such stress-induced cardiac and vascular events \cite{hering2015role}.

Recently, a novel method was proposed for collecting skin sympathetic nerve activity (SKNA) non-invasively through conventional electrocardiogram (ECG) electrodes by analyzing high-frequency components of the signal above $500$ Hz, requiring a sampling frequency beyond $2000$ Hz. Termed NeuECG, this method has prompted several studies investigating the effectiveness of SKNA indices in evaluating SNS activity. For instance, SKNA has been associated with ventricular rate acceleration during atrial fibrillation \cite{4}, and high SKNA levels have been observed in patients with recurrent syncope \cite{5}. Additionally, SKNA has been shown to precede the onset and termination of paroxysmal atrial tachycardia and fibrillation \cite{6}. In patients with overactive bladder, sympathetic activity measured by SKNA was significantly higher compared to healthy controls and decreased significantly after treatment \cite{7}.

In our previous study, we investigated the response of SKNA to cognitive stress induced by performing the Stroop color and word test and showed that an amplitude-based time-domain feature of SKNA, which we called high-amplitude intervals (HaSKNA), can classify cognitive stress from the baseline relaxed mental state \cite{8}. Our protocol was designed to minimize motion and prevent any possible muscle noise artifacts.

However, to effectively study the SNS and the adverse effects of its chronic dysfunction on heart conditions, including those induced by stress, we need continuous data collected over time in real-life scenarios (similar to ECG recording through Holter monitors) where various noise sources, including muscle noise, are present.

A study presented a portable monitoring system for simultaneous non-invasive recording of ECG and SKNA with a sampling rate of $4000$ Hz \cite{9}. The system incorporates adaptive power-line-interference filtering and built-in motion artifact rejection. The system demonstrated a lower noise floor compared to the reference PowerLab device and achieved a high correlation coefficient with the reference SKNA envelope, suggesting a similar effect as NeuECG-based data acquisition \cite{10}. Furthermore, clinical results highlighted the system's ability to reflect sympathetic nerve activity alterations before and after anesthesia injection. Using the same device, \cite{11} suggests that intracerebral hemorrhage patients with reduced average SKNA (aSKNA) could have a worse prognosis.

In this study, we show that muscle noise is a serious interference in SKNA analysis and propose a deep neural network model for detecting 1-second windows of muscle noise from SKNA activation induced by cognitive stress, progressing one step forward towards wearable SKNA with minimal false alarms.

\section{Materials and Methods}

\subsection{Participants and Experimental Setup}

The study included 12 healthy participants aged between 25 and 32 years, comprising 7 females and 5 males. All participants provided informed consent before the commencement of the study. The study protocol was approved by the Institutional Review Board of the University of Connecticut. ECG data were collected using a standard two-lead system, specifically lead II and lead III. Electrode placement followed conventional positioning: electrodes were affixed to the participants' wrists and ankles, while a ground electrode was placed below the right rib cage. ECG signals were recorded via a Bio Amp connected to a PowerLab device, and data acquisition was managed using LabChart Pro 7 software (ADInstrument, Sydney, Australia). The sampling frequency was set to $10$ kHz. Participants remained in a supine position for the entire duration of the study. For each participant, one ECG channel was randomly designated to remain stable, while the other channel recorded data during voluntary movements as dictated by the protocol.

The study protocol comprised five phases, each designed to assess ECG responses under controlled conditions of rest and cognitive stress, with interspersed periods of voluntary muscle movements.

\textbf{Phase 1 – Baseline Recording (Round 1):} Participants maintained a relaxed supine position without any voluntary movements. Data were recorded for a duration of two minutes.

\textbf{Phase 2 – Stroop Task (Round 1):} Participants were subjected to a Stroop color and word test to induce cognitive stress \cite{12}. A smart tablet displayed color-denoted words where the text color often differed from the color name, against differently colored backgrounds (e.g., the word "blue" written in yellow ink with a purple background). Participants were instructed to name the text color. The tablet, positioned on a holder next to the bed, was tilted for easy viewing. During this task, participants were randomly assigned one of six colors and instructed to perform a controlled flexion of one arm associated with the ECG channel capturing movement upon seeing the assigned color. The experiment conductor annotated instances of flexion accurately. This task was conducted for five minutes.

\textbf{Phase 3 – Baseline Recording (Round 2):} Participants returned to the initial relaxed supine position. Following a thirty-second rest period, ECG data were recorded for another two minutes.

\textbf{Phase 4 – Stroop Task (Round 2):} A second round of the Stroop task was conducted using the same procedure as Phase 2. However, this time, a different color was randomly selected for flexion, and participants performed the flexion with their calf muscle associated with the same ECG channel. Phases 2 and 4 were alternated randomly to avoid pattern effects.

\textbf{Phase 5 – Post-task Flexing Session:} Finally, thirty seconds after the completion of the second round of the Stroop task, a two-minute recording of participants in a relaxed supine position was conducted. This time, participants were instructed to perform flexing in their arm and calf muscles when prompted. Five instances of each flexing movement (in arm and calf muscles) were collected.

\begin{figure}[htbp]
\centerline{\includegraphics [width=0.5\textwidth]{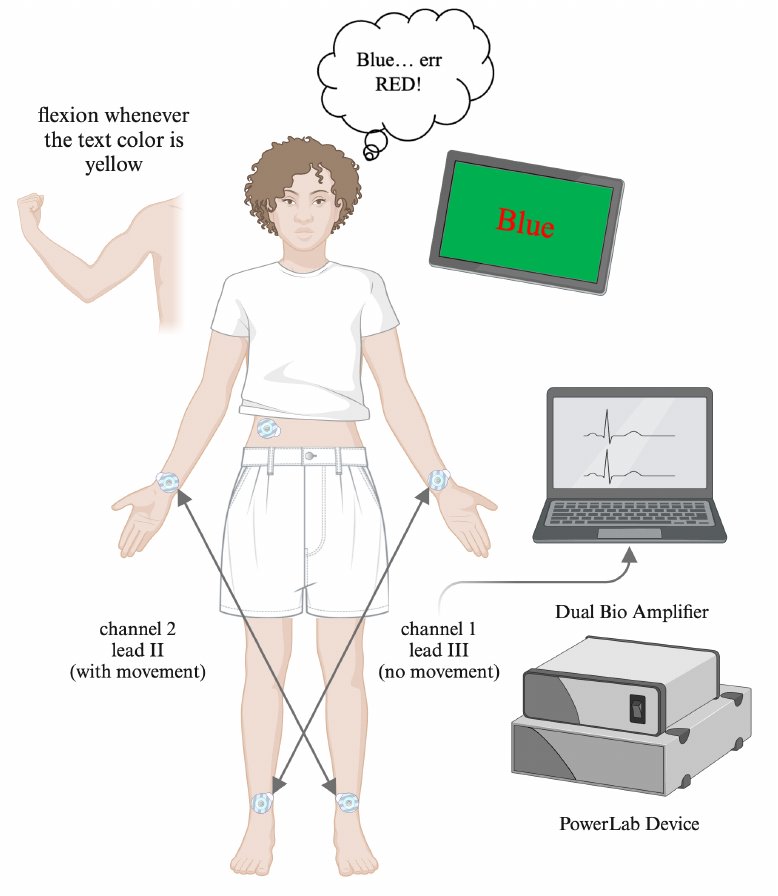}}
    \caption{Experimental Setup. The figure was created with BioRender.com.}
\label{exp_setup}
\end{figure}

Each phase and the flexing events during the experiments were carefully annotated. Figure \ref{exp_setup} shows an overview of the experimental setup. Subsequent analysis was performed using MATLAB R2022 (MathWorks, Natick, MA, USA) and the Python programming language.

\subsection{Preprocessing and Time-frequency Domain Analysis}
In this study, MATLAB was employed to read the \texttt{".adicht"} files generated by LabChart and to generate the dataset. The SKNA signals were extracted by filtering the ECG signals within the frequency range of $500$ to $1000$ Hz. Subsequently, these signals were segmented using a window size of $1$ second. Labels were assigned to each segment to differentiate between baseline periods, Stroop task periods, and Stroop task periods with flexing.

\begin{table}[htbp]
\centering
\caption{An Overview of The Number of Segments in The Dataset}
\renewcommand{\arraystretch}{2}
\small 
\begin{tabular}{c c c c }
\hline
\textbf{}       & \textbf{Baseline} & \textbf{Stroop Task} & \textbf{Stroop + Flexing} \\ \hline
\textbf{Subject 1}  & 245   & 478  & 122 \\
\textbf{Subject 2}  & 243   & 515  & 98  \\
\textbf{Subject 3}  & 242   & 465  & 139 \\
\textbf{Subject 4}  & 243   & 491  & 129 \\
\textbf{Subject 5}  & 243   & 442  & 138 \\ 
\textbf{Subject 6}  & 243   & 488  & 120 \\ 
\textbf{Subject 7}  & 242   & 472  & 135 \\ 
\textbf{Subject 8}  & 244   & 465  & 141 \\ 
\textbf{Subject 9}  & 242   & 454  & 152 \\ 
\textbf{Subject 10} & 254   & 214  & 86  \\ 
\textbf{Subject 11} & 242   & 452  & 146 \\ 
\textbf{Subject 12} & 254   & 457  & 132 \\ \hline
\textbf{Total}  & \textbf{2937} & \textbf{5393} & \textbf{1538} \\ \hline
\end{tabular}
\label{tab:dataset_overview}
\end{table}

To accommodate potential annotation errors and capture a broader window around the flexion events, an annotation error of $0.5$ seconds and a flexion length of 0.5 seconds were defined. Segments exhibiting an overlap greater than $25\%$ with the flexion window were considered to contain flexing events. Finally, the processed datasets for each subject were stored in HDF5 format for subsequent use in Python. Table \ref{tab:dataset_overview} provides an overview of the dataset, detailing the number of segments in each class for each subject.

We used SKNA data from phase 5 associated with the ECG channel, which was subjected to movements, to obtain samples of pure muscle activity at rest (without sympathetic stimulation). Additionally, we utilized segments of the Stroop task without movement, associated with the other channel that remained still and free of muscle interference. We calculated the average power spectral density (PSD) for each of these conditions across subjects in the frequency bands of interest ($500-1000$ Hz) after filtering. Then, we computed the signal-to-muscle noise interference ratio (SMIR) over these bands to assess muscle interference prior to any further attempts to remove it.

We used the created dataset to generate spectrograms for each segment. This was done using a Hann window with a window length of $100$ samples and a hop length of $50$ samples.

\subsection{Classification}
\begin{figure}[htbp]
\centerline{\includegraphics [width=0.35\textwidth]{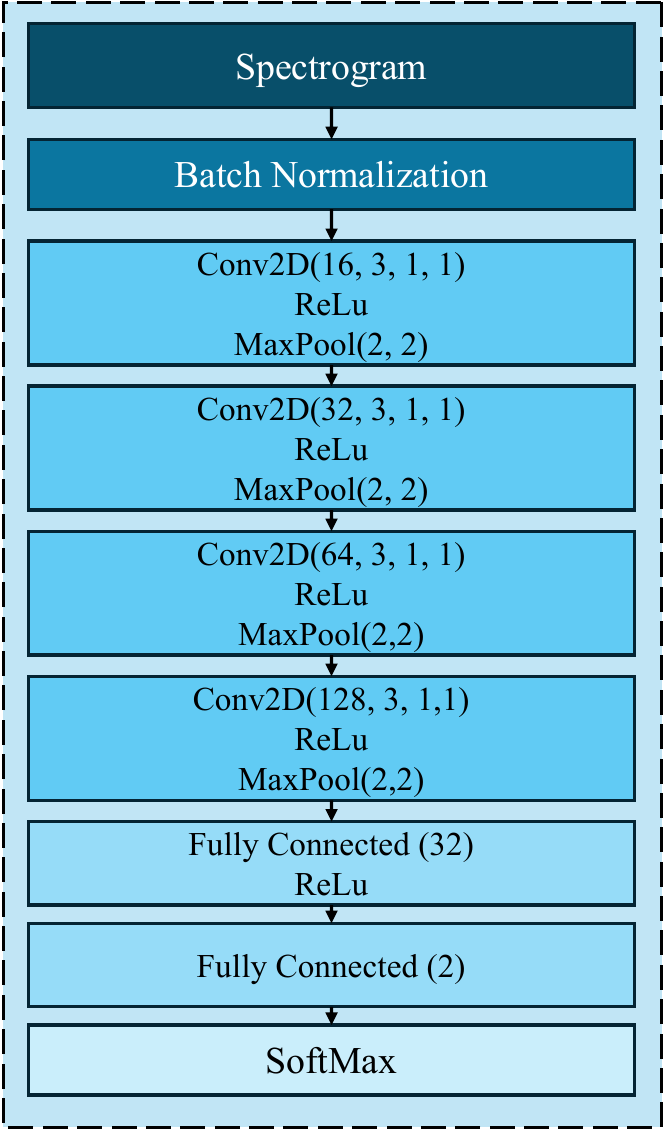}}
    \caption{Network Architecture}
\label{network}
\end{figure}

To classify between baseline periods, Stroop task periods, and Stroop task periods with flexing, we used a 2D deep convolutional neural network model with the spectrograms as inputs. The network architecture is illustrated in Figure \ref{network}.

We utilized the Cross Entropy loss function, with adjusted weights inversely proportional to the frequency of the number of samples in each class to address the imbalance in class distribution. After each layer, except for the last, we applied a dropout with a probability of $0.2$ to prevent overfitting. Prior to training, within-subject normalization was performed. The batch size was set to $32$, and the initial learning rate was $0.001$, using the Adam optimizer.

For testing, we conducted 5-fold cross-validation with subject-specific models, with a test size of $20\%$ of the data. Additionally, we assigned $20\%$ of the training set for validation.

\section{Results}
\begin{figure}[htbp]
\centerline{\includegraphics [width=0.5\textwidth]{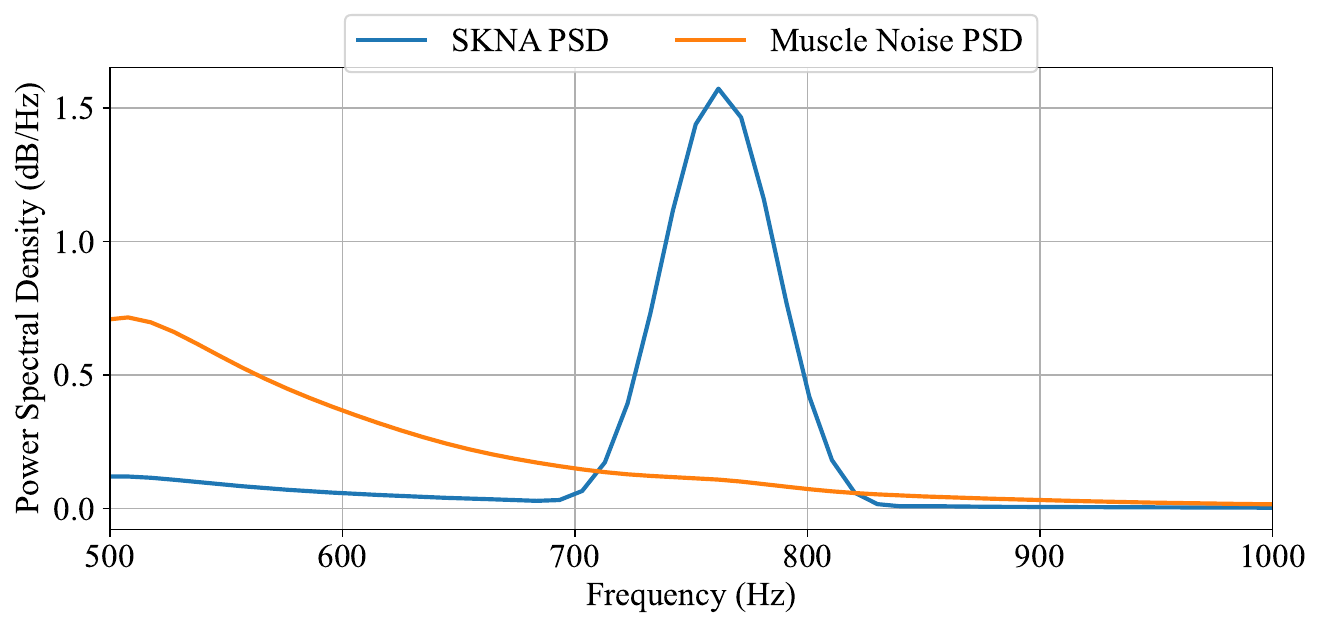}}
    \caption{Average PSD of the SKNA signal during the Stroop task and the muscle noise interference at rest}
\label{PSD}
\end{figure}

Figure \ref{PSD} illustrates the average PSD of the SKNA signal during the Stroop task and the average PSD of muscle noise interference at rest across all subjects. We observe that although the majority of the muscle noise energy is spread over lower frequencies, it still exhibits significant energy within the $500-1000$ Hz frequency band, thereby interfering with the SKNA signal of interest.

\begin{figure}[htbp]
\centerline{\includegraphics [width=0.5\textwidth]{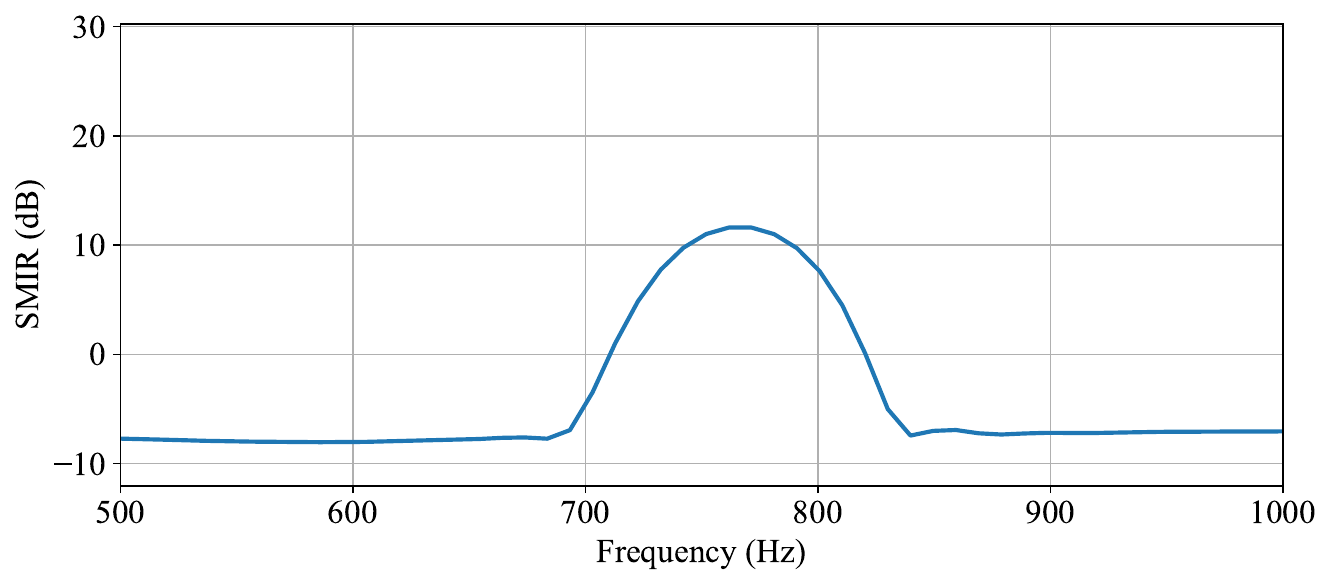}}
    \caption{SMIR over $500-1000$ Hz frequency bands}
\label{SMIR}
\end{figure}

The signal's $95\%$ energy band spans from $527$ Hz to $810$ Hz, while the interference occupies a $95\%$ energy band ranging from $508$ Hz to $898$ Hz. Figure \ref{SMIR} confirms the low SMIR of the signal below $700$ Hz, highlighting the necessity of removing segments with muscle flexing from the analysis to minimize interference and enhance signal clarity.

We ran the personalized models for $200$ epochs, employing checkpoints to save the best-performing model on the validation set within these epochs. The average accuracy across the $5$ folds for each subject, combined across all subjects, was $89.85\%$ for the three classes. The overall normalized confusion matrix, illustrating the model's performance in distinguishing between the baseline periods, Stroop task periods, and Stroop task periods with flexing, is shown in Figure \cite{CM}.

\begin{figure}[htbp]
\centerline{\includegraphics [width=0.45\textwidth]{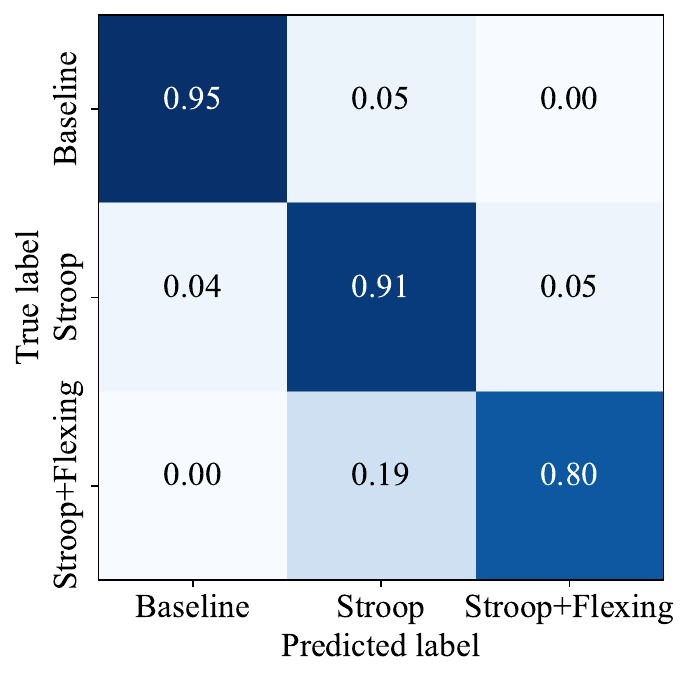}}
    \caption{Overall confusion matrix}
\label{CM}
\end{figure}

\section{DISCUSSION AND FUTURE WORK}
In this study, we demonstrated that despite the common understanding that myopotential interference predominantly occurs below $500$ Hz and thus does not interfere with SKNA \cite{10}, the power of muscle noise within the $500-1000$ Hz range is comparable to that of sympathetic stimulation (SKNA). Without accounting for muscle noise, numerous false alarms can occur. This study aims to provide the initial steps toward the long-term monitoring of the SNS through SKNA, emphasizing the need to account for and remove muscle noise for accurate diagnostics.

This is crucial for the development of wearable SKNA sensors in real-life applications. For hospitalized patients, simply removing the segments with muscle noise may suffice, but for future applications, a method for reconstructing the signals should be developed.

Additionally, in our work, we trained personalized models for each subject; however, developing a general model would be more beneficial.

On another note, accurate data labeling is also essential; we set a window length of $1$ second, an annotation error margin of $0.5$ seconds, and flex lengths of $0.5$ seconds. These parameters may not be optimal and could explain the model’s slight inaccuracies in distinguishing muscle noise.

\bibliographystyle{ieeetr}
\bibliography{ref}

\begin{thebibliography}{10}

\bibitem{sinski2006study}
M.~Sinski, J.~Lewandowski, P.~Abramczyk, K.~Narkiewicz, and Z.~Gaciong, ``Why study sympathetic nervous system,'' {\em J Physiol Pharmacol}, vol.~57, no.~Suppl 11, pp.~79--92, 2006.

\bibitem{floras2009sympathetic}
J.~S. Floras, ``Sympathetic nervous system activation in human heart failure: clinical implications of an updated model,'' {\em Journal of the American College of cardiology}, vol.~54, no.~5, pp.~375--385, 2009.

\bibitem{hering2015role}
D.~Hering, K.~Lachowska, and M.~Schlaich, ``Role of the sympathetic nervous system in stress-mediated cardiovascular disease,'' {\em Current hypertension reports}, vol.~17, pp.~1--9, 2015.

\bibitem{4}
T.~Kusayama, A.~Douglas~II, J.~Wan, A.~Doytchinova, J.~Wong, G.~Mitscher, S.~Straka, C.~Shen, T.~H. Everett~IV, and P.-S. Chen, ``Skin sympathetic nerve activity and ventricular rate control during atrial fibrillation,'' {\em Heart Rhythm}, vol.~17, no.~4, pp.~544--552, 2020.

\bibitem{5}
T.-C. Huang, N.-Y. Chi, C.-S. Lan, C.-J. Chen, S.-J. Jhuo, T.-H. Lin, Y.-H. Liu, L.-F. Chou, C.-W. Chang, W.-S. Liao, {\em et~al.}, ``High skin sympathetic nerve activity in patients with recurrent syncope,'' {\em Journal of Personalized Medicine}, vol.~11, no.~11, p.~1053, 2021.

\bibitem{6}
A.~Uradu, J.~Wan, A.~Doytchinova, K.~C. Wright, A.~Y. Lin, L.~S. Chen, C.~Shen, S.-F. Lin, T.~H. Everett~IV, and P.-S. Chen, ``Skin sympathetic nerve activity precedes the onset and termination of paroxysmal atrial tachycardia and fibrillation,'' {\em Heart rhythm}, vol.~14, no.~7, pp.~964--971, 2017.

\bibitem{7}
Y.-C. Chen, H.-W. Chen, T.-C. Huang, T.-Y. Chu, Y.-S. Juan, C.-Y. Long, H.-Y. Lee, S.-P. Huang, Y.-P. Liu, C.-J. Chen, {\em et~al.}, ``Skin sympathetic nerve activity as a potential biomarker for overactive bladder,'' {\em World Journal of Urology}, vol.~41, no.~5, pp.~1373--1379, 2023.

\bibitem{8}
F.~Baghestani, Y.~Kong, W.~D’Angelo, and K.~H. Chon, ``Analysis of sympathetic responses to cognitive stress and pain through skin sympathetic nerve activity and electrodermal activity,'' {\em Computers in Biology and Medicine}, vol.~170, p.~108070, 2024.

\bibitem{9}
Y.~Xing, Y.~Zhang, C.~Yang, J.~Li, Y.~Li, C.~Cui, J.~Li, H.~Cheng, Y.~Fang, C.~Cai, {\em et~al.}, ``Design and evaluation of an autonomic nerve monitoring system based on skin sympathetic nerve activity,'' {\em Biomedical Signal Processing and Control}, vol.~76, p.~103681, 2022.

\bibitem{10}
T.~Kusayama, J.~Wong, X.~Liu, W.~He, A.~Doytchinova, E.~A. Robinson, D.~E. Adams, L.~S. Chen, S.-F. Lin, K.~Davoren, {\em et~al.}, ``Simultaneous noninvasive recording of electrocardiogram and skin sympathetic nerve activity (neuecg),'' {\em Nature protocols}, vol.~15, no.~5, pp.~1853--1877, 2020.

\bibitem{11}
W.~Wang, H.~Cheng, Y.~Zhang, C.~Cui, Z.~Lin, Y.~Xing, X.~Zhong, X.~Liang, Q.~Cao, Y.~Chen, {\em et~al.}, ``Skin sympathetic nerve activity as a biomarker for outcomes in spontaneous intracerebral hemorrhage,'' {\em Annals of Clinical and Translational Neurology}, vol.~10, no.~7, pp.~1136--1145, 2023.

\bibitem{12}
F.~Scarpina and S.~Tagini, ``The stroop color and word test,'' {\em Frontiers in psychology}, vol.~8, p.~557, 2017.

\end{thebibliography}
\end{document}